\documentclass[letterpaper]{article} 
\usepackage{aaai24}  
\usepackage{times}  
\usepackage{helvet}  
\usepackage{courier}  
\usepackage[hyphens]{url}  
\usepackage{graphicx} 
\usepackage{natbib}  
\usepackage{caption} 
\usepackage{algorithm}
\usepackage{algorithmic}

\usepackage{multirow}
\usepackage{arydshln} 
\usepackage{booktabs}
\usepackage{amsmath}
\newcommand*\samethanks[1][\value{footnote}]{\footnotemark[#1]}
\usepackage{newfloat}
\usepackage{listings}
\DeclareCaptionStyle{ruled}{labelfont=normalfont,labelsep=colon,strut=off} 
\floatstyle{ruled}
\newfloat{listing}{tb}{lst}{}
\floatname{listing}{Listing}
\title{Primitive-based 3D Human-Object Interaction Modelling and Programming}
\author{
Siqi Liu,~~
Yong-Lu Li\thanks{Corresponding authors.},~~
Zhou Fang,~~
Xinpeng Liu,~~
Yang You,~~
Cewu Lu\samethanks.
}
\affiliations{
Shanghai Jiao Tong University

\{magi-yunan, yonglu\_li, joefang, qq456cvb, lucewu\}@sjtu.edu.cn,
xinpengliu0907@gmail.com
}

\let\oldtwocolumn\twocolumn
\renewcommand\twocolumn[1][]{%
\oldtwocolumn[{#1}{
	\begin{center}
		\includegraphics[width=\textwidth]{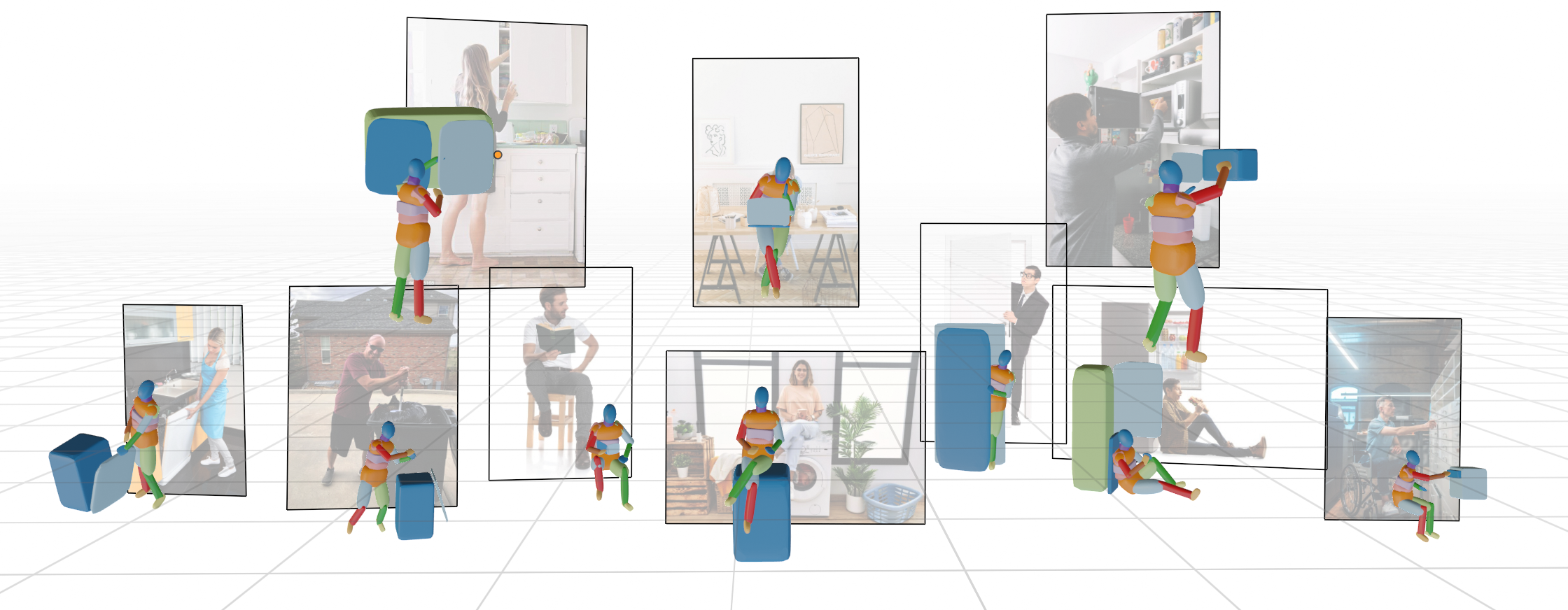}
		\captionof{figure}{We propose a new dataset: P3HAOI, which contains RGB images and the corresponding 3D GT shape of the humans and the objects. All the shapes are represented in primitives.}
		\label{fig:firstpage}
	\end{center}
}]
}

\begin{document}


\maketitle


\begin{abstract}
Embedding Human and Articulated Object Interaction (HAOI) in 3D is an important direction for a deeper human activity understanding. Different from previous works that use parametric and CAD models to represent humans and objects, in this work, we propose a novel 3D geometric primitive-based language to encode both humans and objects. Given our new paradigm, humans and objects are all compositions of primitives
instead of heterogeneous entities. Thus, mutual information learning may be achieved between the limited 3D data of humans and different object categories.
Moreover, considering the simplicity of the expression and the richness of the information it contains, we choose the superquadric as the primitive representation.
To explore an effective embedding of HAOI for the machine, 
we build a new benchmark on 3D HAOI consisting of primitives together with their images and propose a task requiring machines to recover 3D HAOI using primitives from images.
Moreover, we propose a baseline of single-view 3D reconstruction on HAOI.
We believe this primitive-based 3D HAOI representation would pave the way for 3D HAOI studies.
Our code and data are available at \url{https://mvig-rhos.com/p3haoi}. 
\end{abstract}

\section{Introduction}

Human and Object Interaction (HOI) has attracted attention. 
Thanks to the progress in deep learning and the emergence of large-scale datasets, achievements have been reached in HOI in 2D space. However, the 3D field has received little focus due to the scarcity of 3D HOI data.
Recently, significant progress has been made in 3D human and object reconstruction respectively. This inspires us that it is time to bring HOI into 3D which would advance various areas like virtual reality and robotics.

However, there are limited datasets that describe HOI with 3D ground truth (GT)~\cite{xu2021d3dhoi,bhatnagar2022behave, huang2022intercap}. 
BEHAVE \cite{bhatnagar2022behave} and InterCap \cite{huang2022intercap} focus on whole-body humans interacting with rigid objects. They contain RGBD video frames, pseudo GT 3D human and object CAD models, as well as contact labels. 
But in the real world, the \textbf{articulated} objects also play an essential role.
Although D3D-HOI \cite{xu2021d3dhoi} contains articulated objects, they only use simplified CAD models from SAPIEN \cite{xiang2020sapien}, which limits its applications. 
In detail, they have several disadvantages:
\textbf{(1)} In terms of \textbf{objects}, 
scanning or building all object CAD models is unscalable because different categories have different mesh topologies and different instances of the same category have different part ratio scales. As a result, they can only select CAD models that look like the objects in the images to be pseudo GT.
(2) In terms of \textbf{humans}, 
the flexibility of parametric models like SMPL is what cannot be ignored when it comes to unseen human body parts in images. When only some part of the body is essential, learning the other body part parameters is redundant.
\textbf{(3)} From the view of \textbf{HOI}, 
representing humans and objects with heterogeneous models like SMPL and CAD models making it hard to generalize to new HAOIs.

In light of these problems, we propose a new 3D primitive-based ``language'' to represent interacted 3D humans and articulated objects. 
We argue that, in 3D HAOI understanding, \textbf{geometric structure} of the object is closer to the essence than semantic categories. For example, while a refrigerator and a closet belong to different categories, their interactions could be similar. However, even within the same category, interactions can differ significantly, as seen with closets having doors with revolute or prismatic joints.
Besides, more categories can be easily covered when creating new objects via assembly primitives.
Based on these observations, we choose to represent all objects in \textbf{primitives}. 
Our primitive-based language is a useful \textbf{complement} to previous representations like CAD models. 
In fact, CAD models are precise representations but in some cases, they contain too many details. When the overall object structure is the focus, these details are not only expensive but also generalize poorly. If we use primitives to represent humans, we can easily choose and model the interested or seen body parts.
Then we choose the representation based on two principles. 
One is it must be \textbf{sufficiently expressive}, and the other is it can be represented in \textbf{compact low-dimensional representations}. The two principles lead us to the representation called \textbf{superquadric}. Superquadric is a powerful primitive that can represent multiple atomic shapes in a single continuous parameter space.

Given our primitive language, a new 3D HOI dataset called P3HAOI (\textbf{P}rimitive-based \textbf{3}D \textbf{H}uman-\textbf{A}rticulated \textbf{O}bject \textbf{I}nteraction) is built (examples in Fig.~\ref{fig:firstpage}).
It contains RGB images and the corresponding primitive-composed pseudo GT 3D humans and objects. We choose 18 kinds of articulated objects, and each category contains 1,305 instances on average.
Aside from this new dataset, we propose a new task: given a single RGB image, jointly reconstruct the human and the object in the image using 3D superquadric primitives, and provide some baseline solutions. 
We treat the human and the object with the same simple presentation and use the same deformation principle, making the 3D HAOI reconstruction solution elegant.
In experiments, our method achieves decent recovery performance.
However, our new task remains challenging. 
Our primitive language would pave a new way for 3D scene understanding and inspire future work. 

Overall, our contributions are:
(1) We propose a 3D superquadric primitive language to represent 3D HAOI, as a vital step to enable unified 3D scene programming.
(2) We build a new dataset containing 23,494 RGB images with both common 3D models and primitive-based models; 
(3) Accordingly, we propose a new task requiring machines to recover 3D interactive humans and articulated objects with primitives and give basic methods as the future baselines.

\section{Related Work}

{\bf Single-view 3D Reconstruction.} 
There are two main streams. One is to reconstruct using a whole model, the other is to use a primitive-composed model. As for 
the former,
one common method is to fit the input or regress the parameters of a 3D deformable template model \cite{8578514, zuffi2019threed, rueegg2022barc, li2020detailed, dwivedi2022learning, pymaf2021, pymafx2023}, which is usually category-specific. The other is to encode the model into a latent manifold and then decode the latent code into the model \cite{park2019deepsdf, sitzmann2020scene, wu2017learning, wu2023magicpony, huang2023shapeclipper}. Both lack flexibility and are not suitable for manipulation. Considering these problems, primitive-based models are proposed \cite{yao2021discovering, kluger2021cuboids, yavartanoo20213dias, paschalidou2021neural, han2021compositionally, he2021single, yao2022lassie, yao2023artic3d, yao2023hilassie}. Most use rendered 3D models as input \cite{yao2021discovering, yavartanoo20213dias, paschalidou2021neural, han2021compositionally}, but few use realistic photos \cite{kluger2021cuboids, he2021single, yao2022lassie, yao2023artic3d, yao2023hilassie}, among which \cite{yao2022lassie, yao2023artic3d, yao2023hilassie} are most similar to ours. However, they only focus on one animal in one image case, and the input images need to be carefully curated or pre-processed. Skeletons are also needed, making the pre-processing more complex.

{\bf 3D Human-Object Interaction (HOI).}
Over the years, 2D HOI understanding has great progress \cite{zhang2021mining, wu2022mining, liu2022interactiveness, li2019transferable}. 
However, less attention has been paid to 3D HOI, especially to whole-body 3D HOI as lacking of 3D data.
The early method like \cite{zhang2020perceiving} detects humans and objects separately and fits single-view images using SMPL and CAD models, but the spatial arrangement of the human and the object is often incorrect due to the lack of in-the-wild 3D GT 
during training.
To pave the way from 2D to 3D, Bhatnagar~\textit{et. al.}~\cite{bhatnagar2022behave} and Huang~\textit{et. al.}~\cite{huang2022intercap} create new datasets containing RGBD videos and pseudo GT 3D human and rigid object models. 
With the help of BEHAVE, \cite{xie2022chore, wang2022reconstruction, xie2023vistracker} get better performance on single-view HOI 3D reconstruction. 
However, there is still an ignored gap from 2D to 3D, that is humans interacting with articulated objects. To fill the gap, \cite{xu2021d3dhoi} creates a 3D HOI dataset, D3D-HOI, focusing on articulated objects and gives a method for reconstructing human and articulated objects from videos. All of these datasets use SMPL and CAD models to represent humans and objects, leading to poor flexibility and generalizability.

\section{P3HAOI Dataset Construction}

To tackle the main obstacle, the lack of 3D data for HAOI, we create P3HAOI ($\textbf{P}$rimitive-composed \textbf{3}D $\textbf{H}$uman-$\textbf{A}$rticulated-$\textbf{O}$bejct $\textbf{I}$nteraction), a dataset which contains 23,494 images and 18 object categories, as well as the corresponding 3D pseudo GT of humans and objects. It is made up of 9,574 real daily life images and 13,920 synthetic photo-realistic images. 
We collected images from the Internet, absorbed frames from D3D-HOI, and shot some videos by ourselves. In our images, at least half of the object and humans have to be clearly observed.
Details of the object categories are shown in Tab.~\ref{tab:table1}. For each object category, $70\%$ is used as the training set and $30\%$ is used as the testing set.

In the following sections, we will describe in detail how we build the dataset.

\subsection{Preliminary}
\label{sec:preli}
\textbf{Articulated Models.}
Articulated objects are objects composed of more than one rigid part connected by joints and allowing hinge, slider, and screw motions.
When the main structure and the relative poses of the rigid parts are the focus, the trivial components and textures are ignored. We collect simplified CAD models of laptops, refrigerators, microwaves, trashcans, washing machines, ovens, dishwashers, and storage from D3D-HOI. Big-size scissors and doors are collected from the SAPIEN Dataset. Books and drawers are referring to AKB-48~\cite{liu2022akb}. We also simplify the later four classes using the CAD model pre-processing method proposed in D3D-HOI. For the car trunk, car hood, car door, and Murphy bed, we created the CAD models by ourselves. We selected categories that follow the principle: the object can not be too small compared with the volume of the humans, in case it is too hard to detect the shape and poses of the objects in the images.


\textbf{Superquadric.}
Superquadrics are a parametric family of surfaces. With only $11$ parameters, they can represent cubes, ellipsoids, cylinders, spheres, \textit{etc}, in different poses and sizes. The explicit representation of a superquadric in the canonical space can be described as $\textbf{r}(\eta, \omega) = \left[A cos^{\epsilon_1} \eta cos^{\epsilon_2} W \right]$, where $\eta \in [-\pi/2, \pi/2]$, $W = [\omega, \omega, 0]$ and $\omega \in [- \pi, \pi]$. $\left[ \epsilon_{1}, \epsilon_{2} \right]$ describes the shape. $A = \left[ \alpha_{1}, \alpha_{2}, \alpha_{3} \right]$ describes the size. 
To describe the pose, 
we use a vector $\left[ t_{1}, t_{2}, t_{3} \right]$ to describe the translation and a quaternion vector $\left[ q_{1}, q_{2}, q_{3}, q_{4} \right]$ to describe the rotation.

It also has an implicit function. Given a point $x$ in the 3D space, we have
\begin{equation}
f(x; \lambda) = ((\frac{x}{\alpha_{1}})^{\frac{2}{\epsilon_2}} + (\frac{y}{\alpha_{2}})^{\frac{2}{\epsilon_2}}))^{\frac{\epsilon_2}{\epsilon_1}} + (\frac{z}{\alpha_3})^{\frac{2}{\epsilon_1}},
\end{equation}
where $\lambda$ is the intrinsic parameters of a superquadric and in detail, it stands for $[\alpha_1, \alpha_2, \alpha_3, \epsilon_1, \epsilon_2]$. $[\alpha_1, \alpha_2, \alpha_3]$ describes the size and $[\epsilon_1, \epsilon_2]$ describes the shape. 
With this implicit function, we can easily tell if $x$ lies inside or outside the surface of the superquadric. Specifically, if $f(x; \lambda) = 1$, $x$ is on the surface. If $f(x; \lambda) \leq 1$, $x$ is inside and if $f(x; \lambda) \large 1$, $x$ is outside.
We choose superquadric as the representation of primitives as its expression is mathematically simple but informative. Moreover, its parametrization is continuous, making it suitable for deep learning.


\begin{table}[!ht]
\centering
\resizebox{\linewidth}{!}{
	\begin{tabular}{cccccc}
		\hline
		door & refrigerator & microwave & trashcan & cardboard & drawer \\ 
		580 & 2349 & 967 & 537 & 560 & 88 \\ \hline
		car\_trunk & car\_hood & washing machine & oven & dishwasher & storage \\ 
		100 & 98 & 747 & 398 & 370 & 488 \\ \hline
		murphy\_bed & book & car\_door & big\_scissors & laptop & pizza box \\ 
		53 & 600 & 94 & 84 & 885 & 576 \\ \hline
\end{tabular}}
\caption{Number of images for the real dataset.}
\label{tab:table1}
\vspace{-8px}
\end{table}

\subsection{Mesh to Superquadrics}
\label{sec:mesh2super}
The pipeline of converting meshes to superquadrics for humans and objects is the same:
(1) segmenting meshes in canonical space based on the structure, (2) converting the segmentations into superquadrics, and (3) transforming these superquadrics into view space poses. For humans, we segment the SMPL \cite{SMPL:2015} mesh in T-pose based on the human part segmentation vertices mapping file provided by the authors. The finger and toe parts are eliminated for the overall structure simplicity. For objects, the CAD models are segmented by pre-defined joints. Then a network is trained to convert these parts into superquadric-composed shapes. To train the network, firstly a mesh encoder is trained. 
Large quantities of randomly generated superquadrics in different shapes and sizes are put into the network proposed in \cite{DBLP:journals/corr/abs-2006-04325} to get the mesh encoder. Then to train our mesh-superquadric converting network, the pre-trained mesh encoder is frozen. Each part segmentation is first encoded by the mesh encoder, and then a decoder is trained to predict the parameters of the superquadric. 
We sample points on the surfaces of the predicted superquadrics and apply Chamfer loss \cite{DBLP:conf/eccv/Gavrila00} with the GT part meshes. When the network overfits, we obtain 
the superquadric-composed shape in canonical space.

Transforming canonical space superquadric-composed shapes into view space is
different between humans and objects. For humans, ROMP \cite{ROMP} is applied to predict the initial poses. Then based on 2D keypoints predicted by Openpose \cite{cao2017realtime} and Alphapose \cite{alphapose, fang2017rmpe, li2019crowdpose}, EFT \cite{joo2020eft} is utilized to fine-tune the initial poses. Finally, the canonical space superquadrics are transformed based on the fine-tuned joint poses, as we apply the same joint definition as SMPL. For objects from our Internet image collection, the transformation matrix
is recorded while annotating. For objects from the D3D-HOI dataset, the provided GT object deformation information is utilized. More details are in the supplementary.

\subsection{3D Data Annotation}
\label{sec:3danno}
\begin{figure*}
\centering
\includegraphics[width=\textwidth]{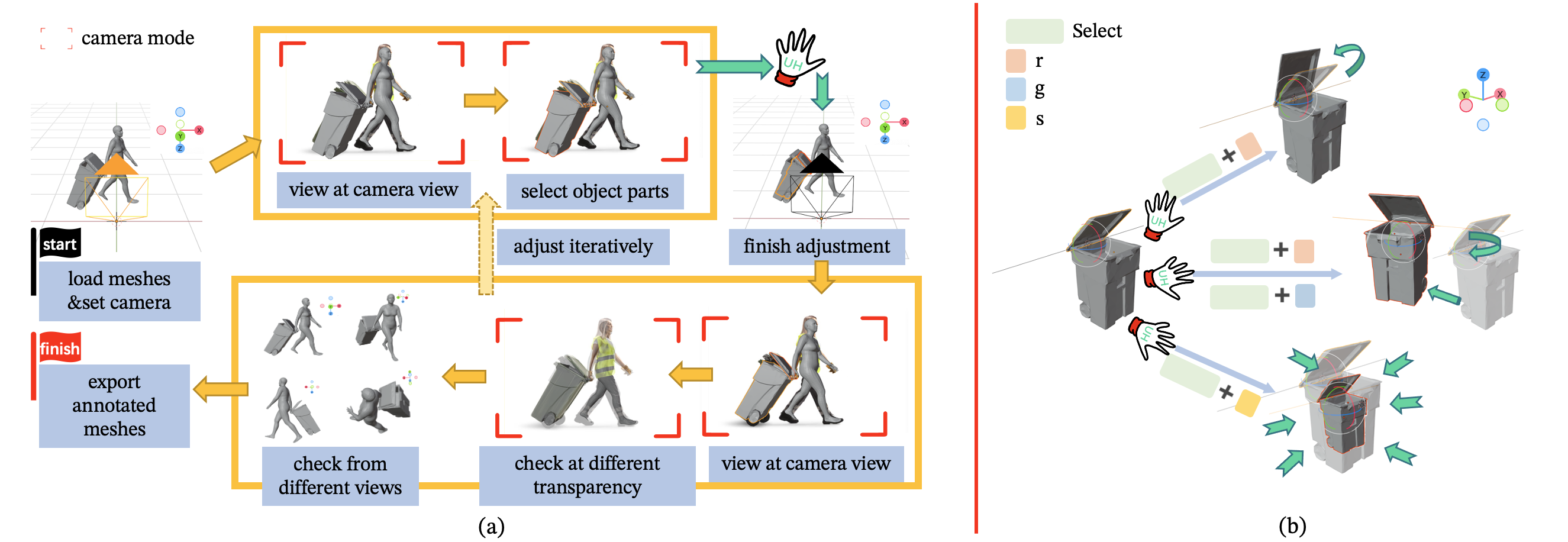}
\vspace{-10px}
\caption{
	We adjust the orientation, location, and scale of objects in Blender. 
	(a) The annotation process.  (b) Details about deforming objects in Blender using UH.
}
\label{fig:anno_1}
\vspace{-10px}
\end{figure*}

We annotate the 3D poses and locations of humans and objects in Blender.
Unlike D3D-HOI, we did not use numerical values to label as people are not sensitive to numerical values regarding location and orientation. Instead, annotating objects in Blender is more intuitive. By setting 
axles for the articulated objects to rotate around, several operations can be operated on the objects, as shown in Fig. \ref{fig:anno_1}.
For simplicity, we name this series of operations the UltraHand (UH).

For each category, we select CAD models with the closest appearance to the object in the image. 
Textures are not taken into consideration when choosing CAD models. 
For each image instance, the fine-tuned SMPL mesh and the object CAD template model are imported into Blender. Then we set the camera parameters predicted by ROMP and the original image as the background image in the camera view mode. During the annotation process, we did not deform the human. All operations take place on the object. Using the UH, we first adjust the object template to the object state shown in the image. Then we switch between the camera view mode and the whole world view mode to check (1) if the annotated object projection aligns perfectly with the background image, and (2) if the spatial arrangement with the human is reasonable based on the information in the image, and (3) if it is consistent with common sense, while continue adjusting the object using the UH. The annotation for one image is finished until the object satisfies the three conditions above. When the annotation is finished, we export the object parts. The whole annotation process is shown in Fig. \ref{fig:anno_1}.
After the experts' annotation, we transfer both SMPL and CAD object models into superquadrics.
Some visualization of our dataset is in Fig.~\ref{fig:firstpage}. 



\subsection{Synthetic Data}
We also generate a large amount of synthetic data.
32 human models from THuman2.0 and 15 from Mixamo are chosen. Then a variety kind of human poses and 3D HAOI scenes can be created manually.
For each scene, we render the scenes from different camera views, as well as individual silhouettes. We also set up various backgrounds from simple to complex, from indoors to the wild. 
Some examples are shown in the Supplementary. 
After rendering, we convert these CAD models into superquadric-composed shapes.


\section{Primitive-based HAOI 3D Reconstruction}


Our goal is to jointly reconstruct humans and the interacted articulated objects from a single-view RGB image using superquadrics.
We will first state some main challenges and then briefly describe possible solutions. Then considering these challenges, we will state why and what solution we choose.

There are three significant challenges. Firstly, it is an inherently ill-posed task.
Secondly, the GT 3D HAOI data is not sufficient.
Thirdly, compared to rigid objects, larger pose space and shape space of articulated objects cause much harder 6-DoF predictions.

\begin{figure*}
\centering
\includegraphics[width=\textwidth]{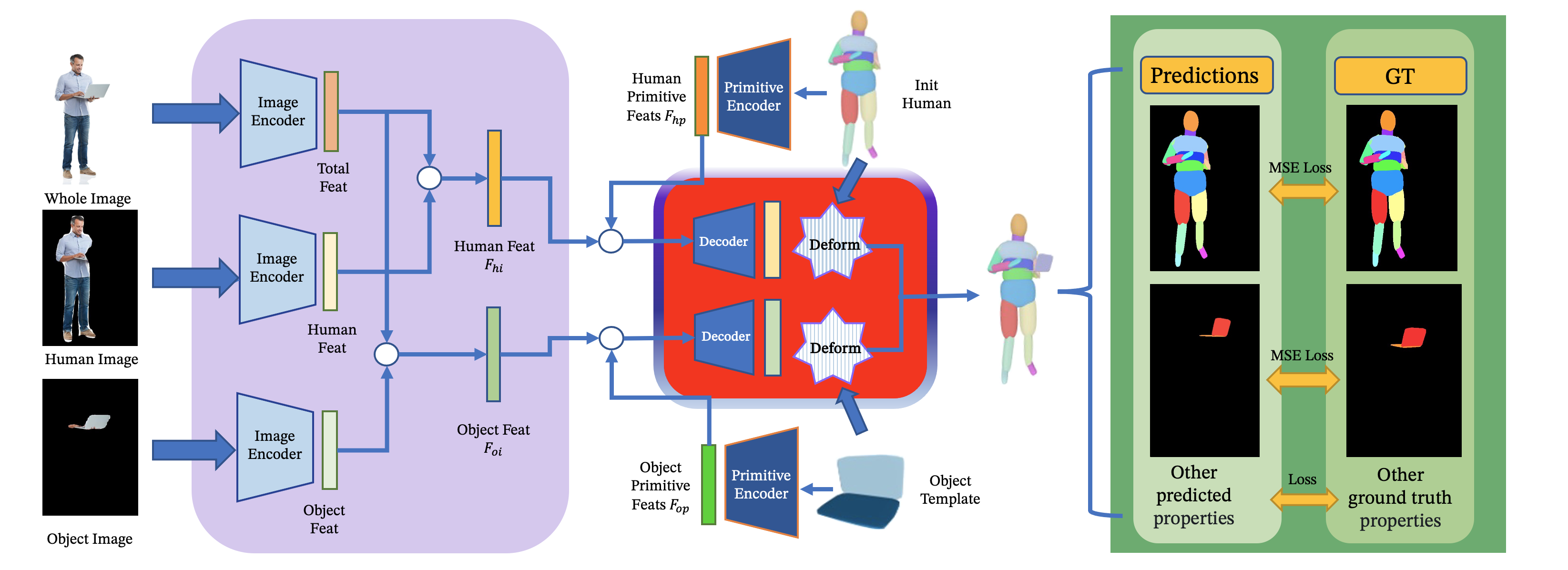}
\vspace{-20px}
\caption{Overview of our single-view 3D HAOI reconstruction method. In the purple block are the image encoders. Superquadric-composed shapes are also encoded to predict primitive features. Then the image features and primitive features are contacted together to predict the 6-Dof of each input superquadric part. Deformation is based on these predictions. The deformed shapes are then rendered back to 2D and some losses are computed based on some supervisions.}
\label{fig:baseline_0}
\vspace{-15px}
\end{figure*}

There are mainly two kinds of solutions. One is to first predict the whole scene and then fit the human and the object models to the lifted scene. The other is to reconstruct the human and the object separately and then combine them based on their spatial arrangement and relative scales. While the former could more easily obtain information about the spatial relationship
, the lifted data are often more noisy, making it more complex and difficult at early fitting steps. 
As for the later method, 
it can be divided into three sub-methods.
The first one is the most straightforward. 
Well-posed human and object models are given at first and only the 6-DoF predictions are needed.
This makes sense when it comes to rigid objects, just as what \cite{zhang2020perceiving} does. 
However, it is unrealistic for articulated objects as there are various part pose states.
So a deeper solution is proposed, that is the human model is prepared, but the poses of the object parts are predicted.
The deepest one is the human part and the object part poses are all predicted. This is the most ideal solution because when people interact with some object, we cannot view the human as a single individual. 
The state of the humans and the objects affect each other.
Thus, joint optimization cannot be ignored if we want a good HAOI 3D reconstruction. 

This paper proposes a compromise between the second and the third methods. 
We first prepare the human pose using a pre-trained human shape recovery model
, then predict the object states.
After the coarse reconstruction
, the information between the human and the object is leveraged for fine-grid optimization.
The method flow chart is shown in Fig.~\ref{fig:baseline_0}. We reconstruct the human and the object separately and then do joint optimization between them. The inputs are RGB images and superquadric-composed templates. Then the extracted features of these inputs are fed into a decoder to predict the scale, rotation, and translation parameters to transform the templates into the view space in the images.
Finally, spatial information between the human and the object is used to do joint optimization to get a better reconstruction.


\textbf{Input Encoding. }
The human image is generated by first detecting the human/object silhouette using Detectron2 \cite{wu2019detectron2}. For object categories within the Detectron2 class list, we only use detections whose scores are at least $0.8$, and for whose scores are less than $0.8$, we annotate the object masks manually. Then other image regions are masked by the silhouettes. DINOv2 \cite{oquab2023dinov2} is applied to extract features from these images. In our experiment, DINOv2 is frozen. We add CNN layers after the DINOv2 feature extractor to fine-tune the predicted features.

For the superquadric-composed templates, points are sampled uniformly on the surface of each primitive. Then the pre-trained mesh encoder in Sec. ~\ref{sec:mesh2super} is applied to predict the features of each superquadric. The parameters of the mesh encoder are frozen during training and testing.
These primitive features are then contacted together to represent the template feature. 


\textbf{Reconstruction of Human. }
Experiments show that regressing different human poses from just a primitive-composed T-pose template is very difficult and thus
a good initialization
is essential. So we use ROMP~\cite{sun2021monocular} to predict the human pose in SMPL and then transform the SMPL model into the superquadric-composed model, which serves as our initialization model to the network. Then residual pose parameters are predicted to adjust the human pose.

\textbf{Reconstruction of Articulated Object. }
Unlike humans, each object category shares the same template in canonical space as the initialization.
We define the object structure as trees. 
For each articulated object, there is one root part and several leaf parts (mostly, there is one leaf part, which means there are two parts in total). For the root part, we only predict its global rotation, translation, and scale. For the leaf parts, we predict the local rotations and scales. The axis around which the leaf parts rotate is predefined together with the input template. 
In deformation, this predefined axis is deformed following the deformation of the root part.


\textbf{Objectives. }
While doing separate reconstruction, the human and the object have similar objectives. To avoid redundancy of expression, we express these objectives uniformly and state the existing differences. 
We only describe objectives for the separate reconstruction in \textbf{Objectives}. Objectives for joint optimization are described in \textbf{Joint Optimization}.


\textit{Part segment masks loss.}
Primitives are rendered by Nvdiffrast \cite{Laine2020diffrast} to get the part segment masks. 
MSE Loss is calculated.

\textit{3D keypoint loss.} 
We compute L1 loss between the GT 3D keypoints and the predictions.

\textit{Joint angle loss.}
This objective means joint pose loss for humans, and MSE Loss is applied.

For objects, ``joint angle'' is defined as the angle between the root and leaf parts, and cosine similarity is applied.

\textit{Surface vertices loss.} 

Chamfer distances are calculated between the uniformly sampled vertices on the surfaces of the GT superquadric-composed shapes and the predictions.

\textit{Auxiliary object IUV loss.}
Inspired by \cite{pymaf2021}, we apply an auxiliary IUV prediction to fine-tune the image features for objects. The GT IUV map is easy to generate based on the properties of our superquadric-based representation: the part index defines the I-map, and ($\eta$, $\omega$) from the superquadric explicit representation defines the UV-map. We render the IUV information in 3D space to 2D images to form the GT IUV map. 
A cross-entropy loss is applied for the I-map. The L1 loss is applied to the UV map.

\textbf{Joint Optimization. }
\label{sec::joint_opt}
In 3D HAOI reconstruction, mutual information needs to be used to get the perfect reconstruction as the pose of people and objects are mutually influential. Early works like \cite{zhang2020perceiving} and \cite{xie2022chore} have shown the importance of contact information. However, in our settings, the human may not closely contact the object. We also consider contacts when the human's position is a little far from the object, \textit{e.g.}, a man watching his laptop. In these situations, it is not easy to define the contact region. Besides, labeling contact vertices is time-consuming. Thus we propose a new objective: \textbf{HOI loss}. 
Our training process is composed of $3$ steps. $(1)$ the object network is frozen and we train for the human. $(2)$ the human network is frozen and we train the object. $3$ both the human network and the object network are fine-tuned, and \textbf{in this step, we do joint optimization}. 
During training, the three steps are continuous without interruption. 
We determine at which epoch to proceed to the next step based on whether the corresponding losses in each stage have been stable. 
We set the weights of HOI loss and interpenetration loss from $0.$ to $1.$ not until $(3)$ starts.
In Steps One and Two, the weights of HOI loss and interpenetration loss are set to $0$. Once starting Step Three, both the two weights are set to $1.$ to train the network for joint optimization.

For the HOI loss, we calculate a vector from each human keypoint to each object keypoint. 
Object keypoints $\textbf{P}_o$ are the centers of each part. 
Human keypoints $\textbf{P}_h$ are what are defined as SMPL. 
Then the HOI vector is defined as $\textbf{v}_{ij} = (\textbf{P}_{hi} - \textbf{P}_{oj})$
and the HOI loss $L_{HOI}$ is defined as
\begin{equation}
\begin{aligned}
L_{HOI} &= \frac{1}{K} \sum_{ij}^K cos(\textbf{v}_{ij}, \hat{\textbf{v}}_{ij}), \\
& i \in \{1, ..., M\}, j \in \{1, ..., N\},
\end{aligned}
\end{equation}
where $M$ and $N$ are the numbers of human keypoints and object keypoints respectively. $K = M \times N$.
$cos$ stands for the cosine similarity. By computing the HOI loss,
better spatial results are obtained. Intuitively, it is like a person pulling a puppet to the right pose and position. Besides HOI loss, we also calculate \textbf{interpenetration loss} inspired by \cite{DBLP:journals/corr/abs-2006-08586}. One difference between ours and theirs is that we do not need to first voxelize the shape and then calculate SDFs. As superquadrics themselves have implicit functions, it is easy to calculate the distances between points at the object's primitive surface to the surface of the human primitives. So fewer computing resources and time are needed.

\begin{figure}
\centering
\includegraphics[width=0.5\textwidth]{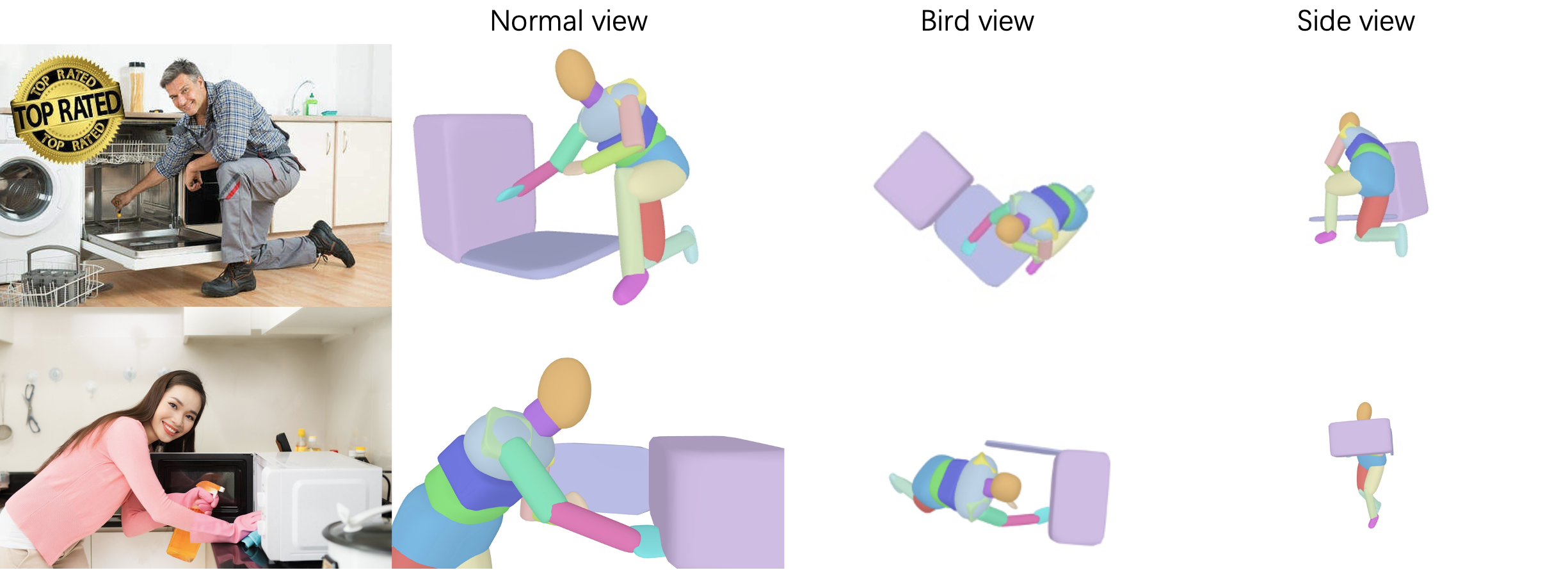}
\vspace{-10px}
\caption{Visualized results. 
}
\label{fig:quali_res}
\vspace{-5px}
\end{figure}

\begin{table}[h!]
\centering
\resizebox{0.5\linewidth}{!}{
\begin{tabular}{c:cc} 
	\toprule[3pt]
	Methods & Human$\downarrow$ & Object$\downarrow$ \\ \hline
	CHORE & 33.7820 & 81.8080 \\ 
	D3D-HOI & 2.0135 & 55.3536 \\ \hline
	Ours & \textbf{0.1241} & \textbf{21.4340}\\ \hline
	\end{tabular}}
	\caption{Quantitative results for baselines. }
	\label{tab:quan_res}
	\vspace{-10px}
\end{table}

\section{Experiments}

\subsection{Baselines}
As reconstructing primitive-composed 3D HAOI is a new task, there are no direct former baselines.
So we modify two methods as the baselines: CHORE~\cite{xie2022chore} and D3D-HOI~\cite{xu2021d3dhoi}, which are closest to our work.

CHORE~\cite{xie2022chore} reconstructs humans and rigid objects from a single RGB image. They select CAD models most similar to the object in the images and use SMPL to represent the human. However, unlike rigid objects, the solution space for articulated objects is more complex and it is unrealistic to prepare objects in one category of all kinds of part states. So in our experiment, for each category, we prepare a CAD model with the most commonly seen part state and view it as a rigid object.
Then we apply the method proposed in CHORE to reconstruct the human and the object. After the reconstruction, the SMPL meshes and object meshes are converted into superquadrics.

D3D-HOI~\cite{xu2021d3dhoi} reconstructs humans and articulated objects from camera-fixed videos, and one model is trained for one video.
They also select the most similar CAD models to the objects in images and predict the 6-DoF of each object part. 
To be fair, we apply D3D-HOI to single-view reconstruction and so we train one model for multiple video frames. All the losses they set can be directly applied to single-view reconstruction, except the contact loss. The contact loss is based on the assumption that the object motions tend to follow the human hand during the interaction and they select representative object vertices to calculate the loss. However, the word ``representative'' means ``moving beyond a threshold during the interaction'', so this concept does not exist in single-view images. 
Following D3D-HOI, humans are predicted directly using EFT, and no fine-tuning is conducted. 
We apply D3D-HOI to reconstruct humans and objects in SMPL and CAD mesh parts. After reconstruction, we convert the reconstrued SMPL mesh and the object mesh into superquadrics to compare with our method. 
More setting details are provided in the supplementary.

\begin{table}[h!]
\centering
\resizebox{\linewidth}{!}{
\begin{tabular}{c:c|cccccc}  
	\toprule[3pt]
	\multicolumn{2}{c}{-}
	&
	\begin{minipage}[b]{0.12\columnwidth}
		\centering
		\raisebox{-.5\height}{\includegraphics[width=\linewidth]{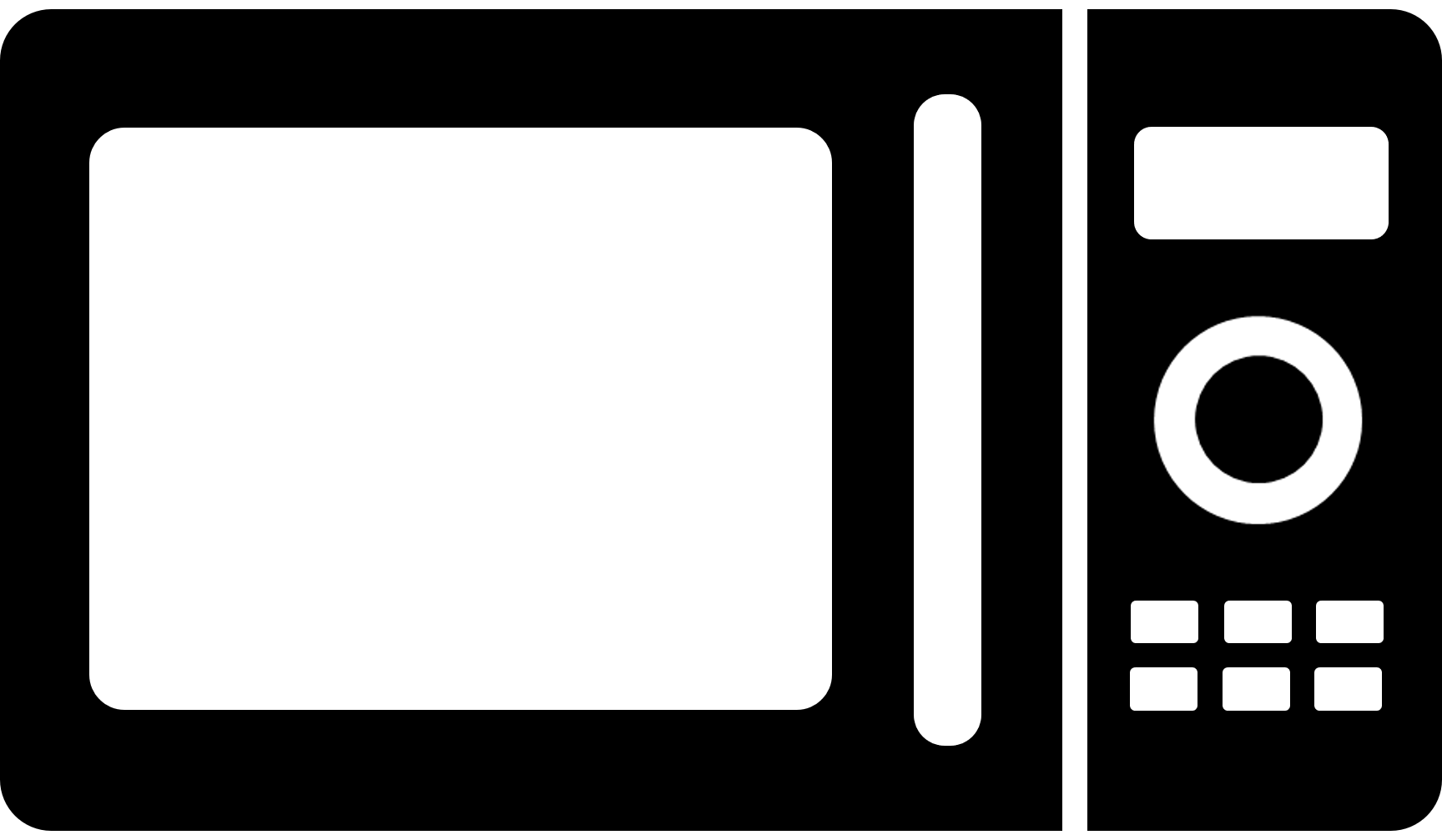}}
	\end{minipage}
	&
	\begin{minipage}[b]{0.12\columnwidth}
		\centering
		\raisebox{-.5\height}{\includegraphics[width=\linewidth]{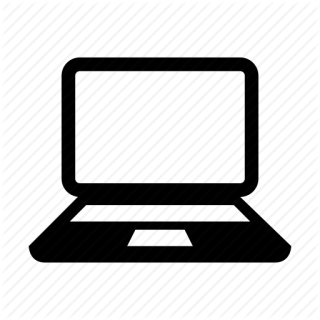}}
	\end{minipage}
	&
	\begin{minipage}[b]{0.11\columnwidth}
		\centering
		\raisebox{-.5\height}{\includegraphics[width=\linewidth]{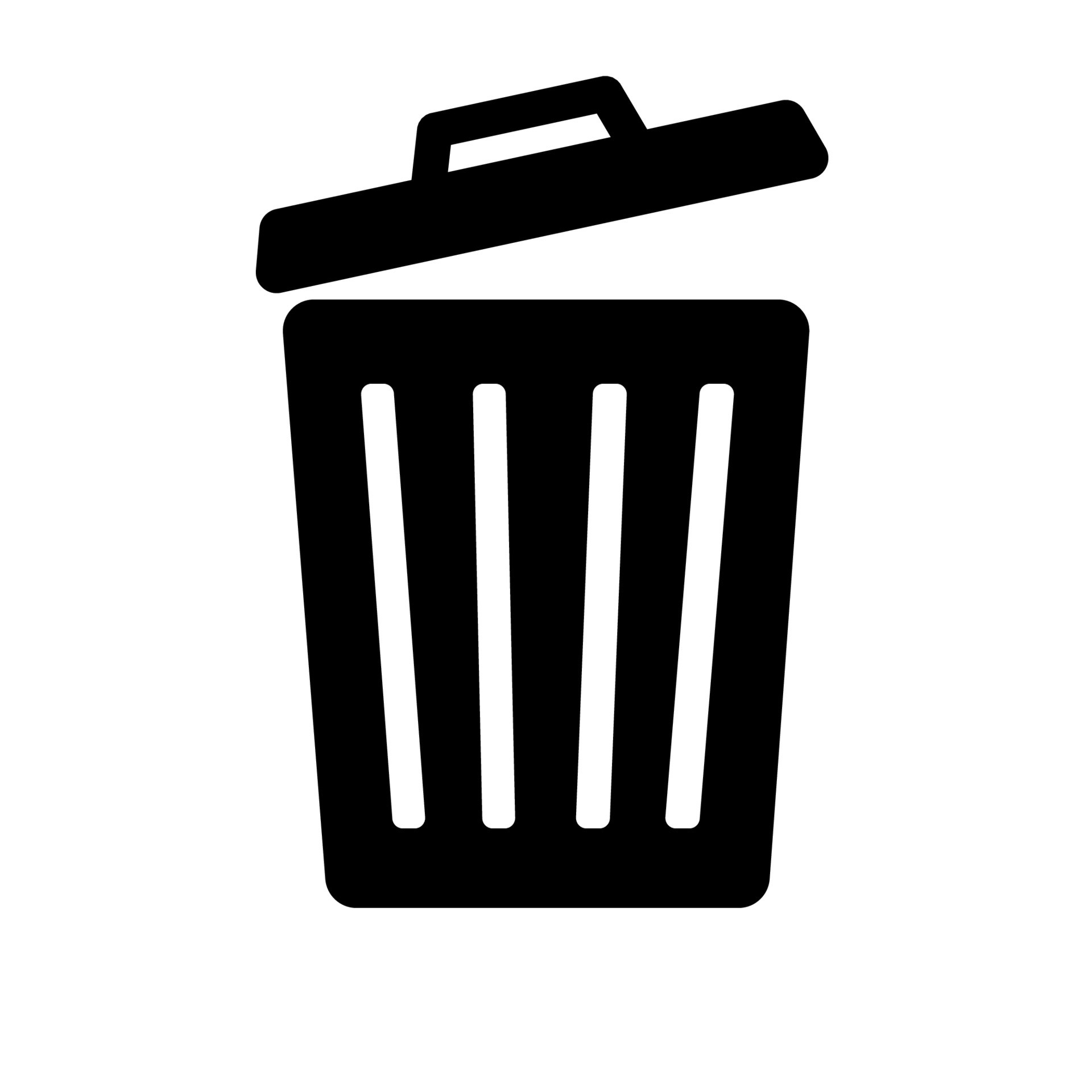}}
	\end{minipage}
	&
	\begin{minipage}[b]{0.17\columnwidth}
		\centering
		\raisebox{-.5\height}{\includegraphics[width=\linewidth]{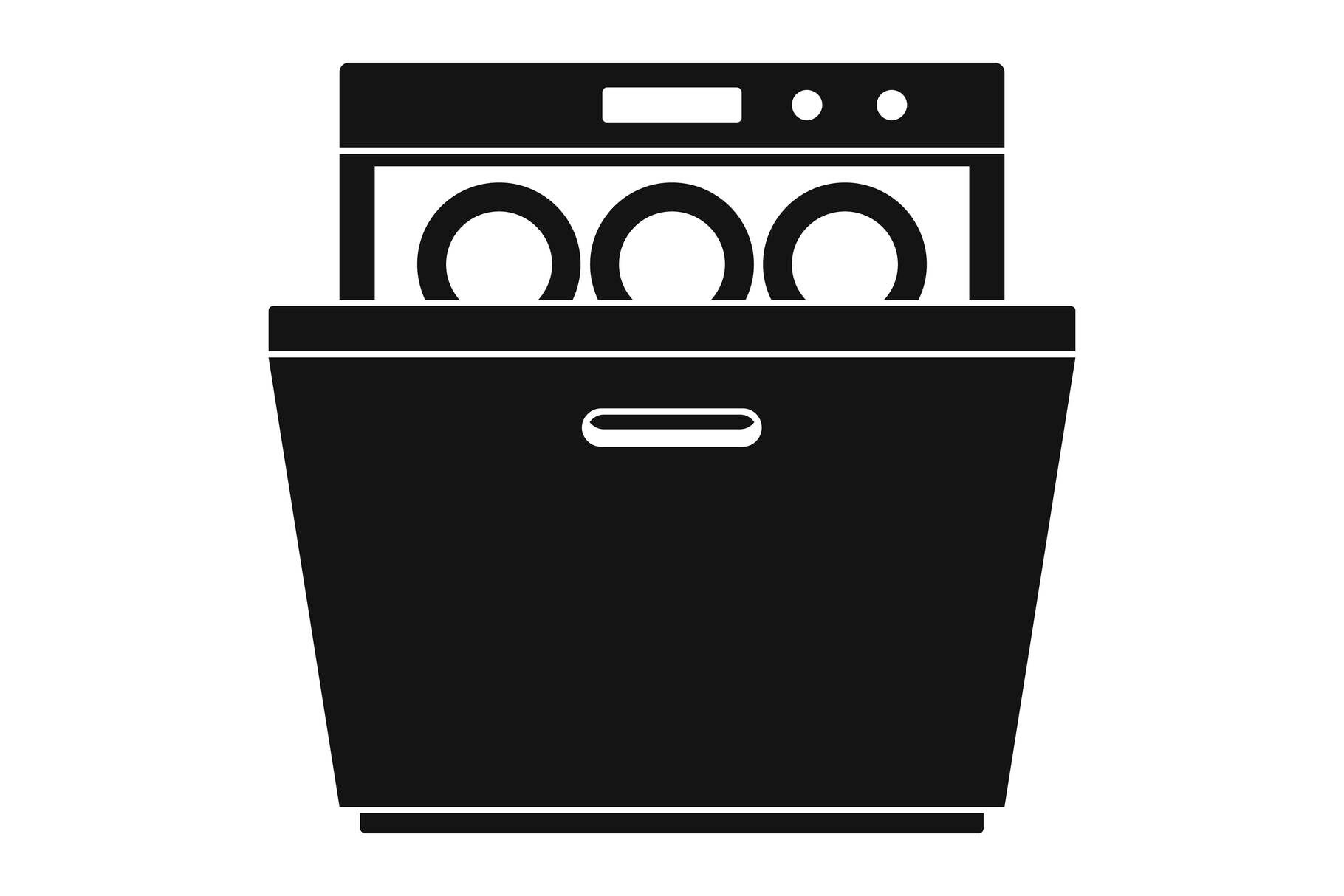}}
	\end{minipage}
	&
	\begin{minipage}[b]{0.1\columnwidth}
		\centering
		\raisebox{-.5\height}{\includegraphics[width=\linewidth]{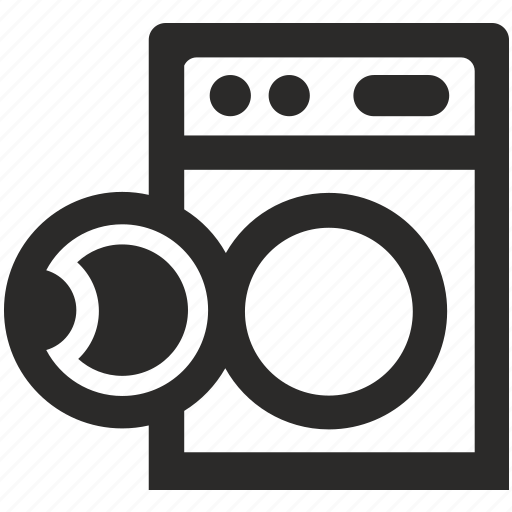}}
	\end{minipage}
	\\ \hline
	\multirow{5}{*}{\textit{H}$\downarrow$} & DINOv1 & 0.4139 & 1.6085 & 0.1773 & 0.3056 & 0.1812 &\\ 
	&DINOv2& 0.0428 & 0.1251 & 0.2234 & 0.1121& 0.1172 &\\ 
	&no IUV& 0.0435 & 0.3418 & 1.1885 & 0.1287 & 0.1146 &\\ 
	&no HOI& 0.0469 & 0.4160 & 0.3462 & 0.1004 & 1.3869 &\\
	&objkp2d& 0.0424 & 0.3428 & 0.9857 & 0.1021 & 0.4452 &\\ 
	\hdashline
	\multirow{5}{*}{\textit{O}$\downarrow$} & DINOv1 & 13.1354 & 16.1319  & 90.3916 & 22.6262 & 14.6895 &\\ 
	&DINOv2& 14.0521 & 8.9078 & 58.0723 & 10.8378 & 15.2990 &\\ 
	&no IUV& 12.8346 & 9.1255 & 32.4188 & 11.9067 & 16.3921 &\\ 
	&no HOI& 12.7995 & 15.2272 & 33.7624 & 13.2436 & 13.3826 &\\ 
	&objkp2d& 13.5021 & 12.8052 & 73.7526 & 11.2008 & 14.1512 &\\ 
	\hline
\end{tabular}
}
\vspace{-10px}
\caption{Quantitative results of the ablation study. }
\label{tab:ablation}
\vspace{-10px}
\end{table}

\begin{figure}
\centering
\includegraphics[width=0.5\textwidth]{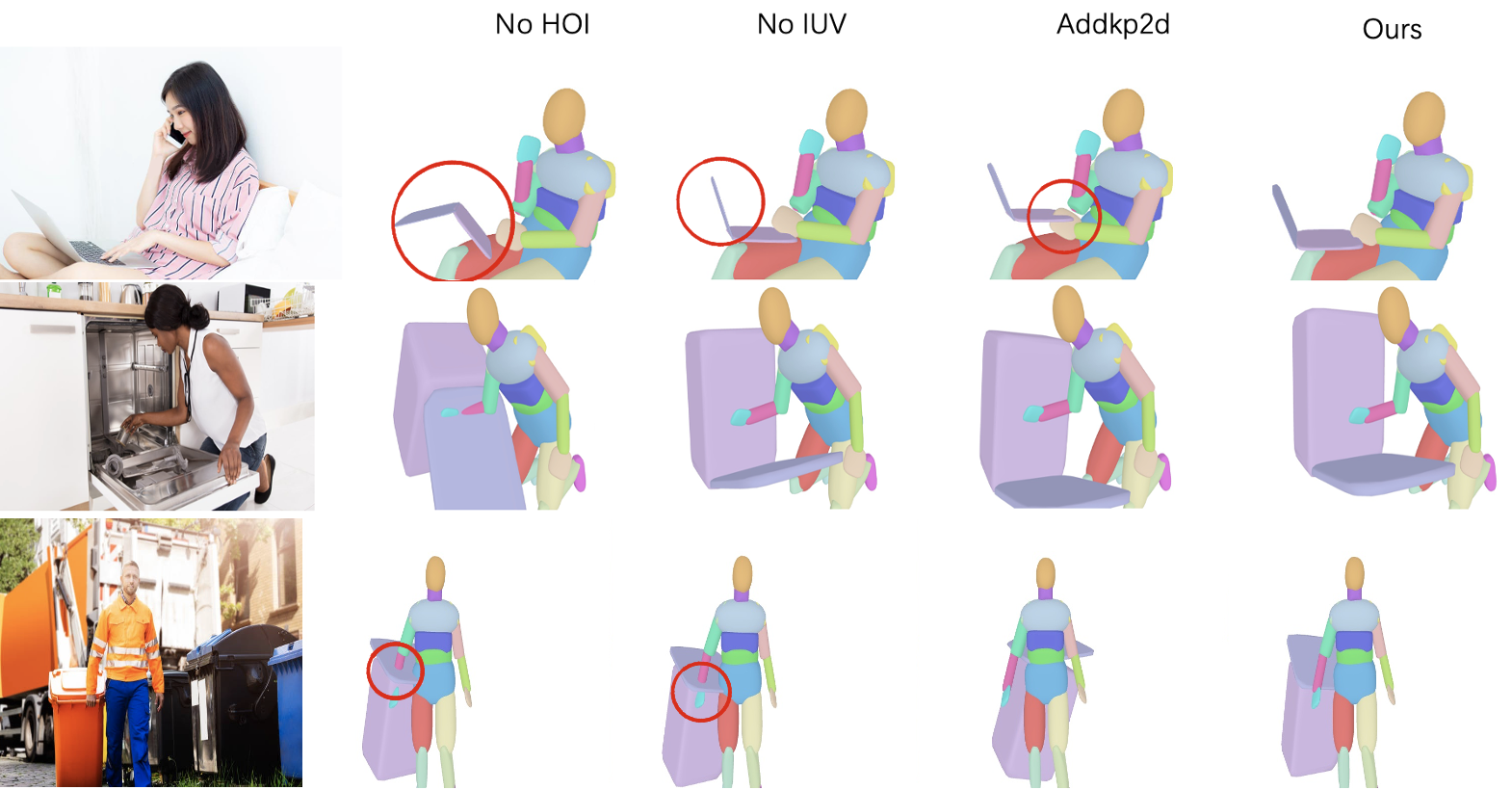}
\vspace{-10px}
\caption{Qualitative ablation results. 
}
\label{fig:abla_res}
\vspace{-15px}
\end{figure}

\subsection{Evaluation}
\textbf{Qualitative Results. }
We evaluate the baselines on three types of P3HAOI datasets: the real dataset, the synthetic dataset, and the fused dataset. Chamfer distances are calculated for humans and objects separately as the evaluation metric. Some qualitative results are shown in Fig. \ref{fig:quali_res}. 
More qualitative results are available in the supplementary.

\textbf{Quantitative Results. }
Similar to previous works \cite{xie2022chore, xu2021d3dhoi}, Chamfer distance is measured in centimeters. To compare, we set the human height to 175cm, the same in D3D-HOI. The quantitative results of the mean Chamfer distance (cm) are shown in Tab. \ref{tab:quan_res}. 
Results show that CHORE fails to handle articulate object reconstruction as they need to train a CHORE field first. However, the shapes of articulated objects are much more complex than rigid objects, so it is more difficult to train such a field. 
For D3D-HOI, due to the lack of contact loss, it is hard for the network to converge, not to mention learn the accurate part-state and global poses of articulated objects. 
Our method captures more information in images for articulated objects compared to D3D-HOI. While the 2D mask term works for rigid objects, it's insufficient for articulated ones. Our part segment mask loss and IUV loss help us capture diverse solution space. Unlike D3D-HOI, our HOI term doesn't force humans to always face objects, allowing for more flexibility and handling various interaction scenarios in daily life, like a person sitting on a washing machine.

\begin{figure}
\centering
\includegraphics[width=0.3\textwidth]{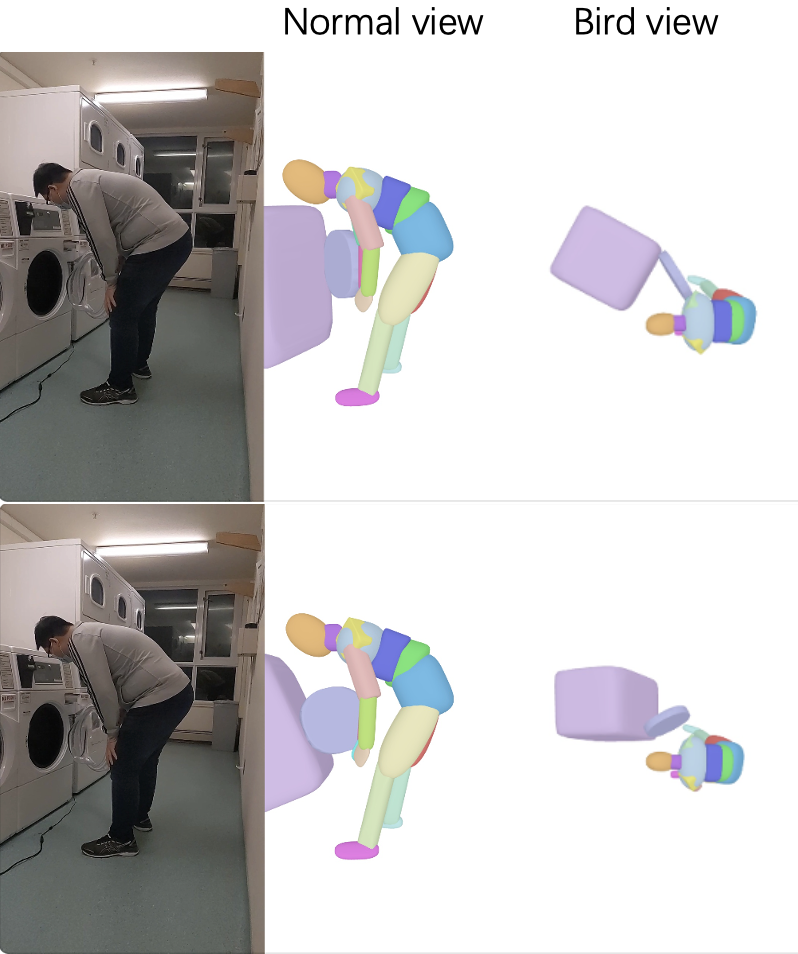}
\vspace{-5px}
\caption{Object 2D keypoint supervision could be disruptions. The 1st and 2nd rows show the result of the same image w/o and w/ 2D keypoint supervision respectively.
}
\label{fig:objkp2d}
\vspace{-10px}
\end{figure}

\textbf{Ablations. }
In Tab.~\ref{tab:ablation}, we ablate our proposed method to analyze some essential components. \textit{DINOv1} and \textit{DINOv2} changes the image encoder backbone. For \textit{DINOv1}, we use the extracted \textit{Key} feature as the image feature. For \textit{DINOv2}, the final token is used as the image feature. \textit{no IUV} does not train the IUV map to fine-tune the image feature extracted by the image encoder backbone. \textit{no HOI} ignores the relationship between the human and the object, which means just reconstructing them separately. \textit{objkp2d} add the object 2D keypoint loss back to prove that it is actually a distraction. 
Some ablation study qualitative results are in Fig. \ref{fig:abla_res}. Results show that the HOI Loss plays an important role in the prediction of the overall pose of the object. Without the HOI Loss, the objects often flip around some axis. The IUV supervision gives more accurate part-state predictions. The disruptions caused by the object 2D keypoint loss are shown in Fig. \ref{fig:abla_res} and Fig. \ref{fig:objkp2d}. The small number of the 2D object keypoints and the inherent ambiguity cause not only inaccurate locations but also rotations.
We provide more visualizations and discussions in the supplementary.


\begin{figure}
\centering
\includegraphics[width=0.8\linewidth]{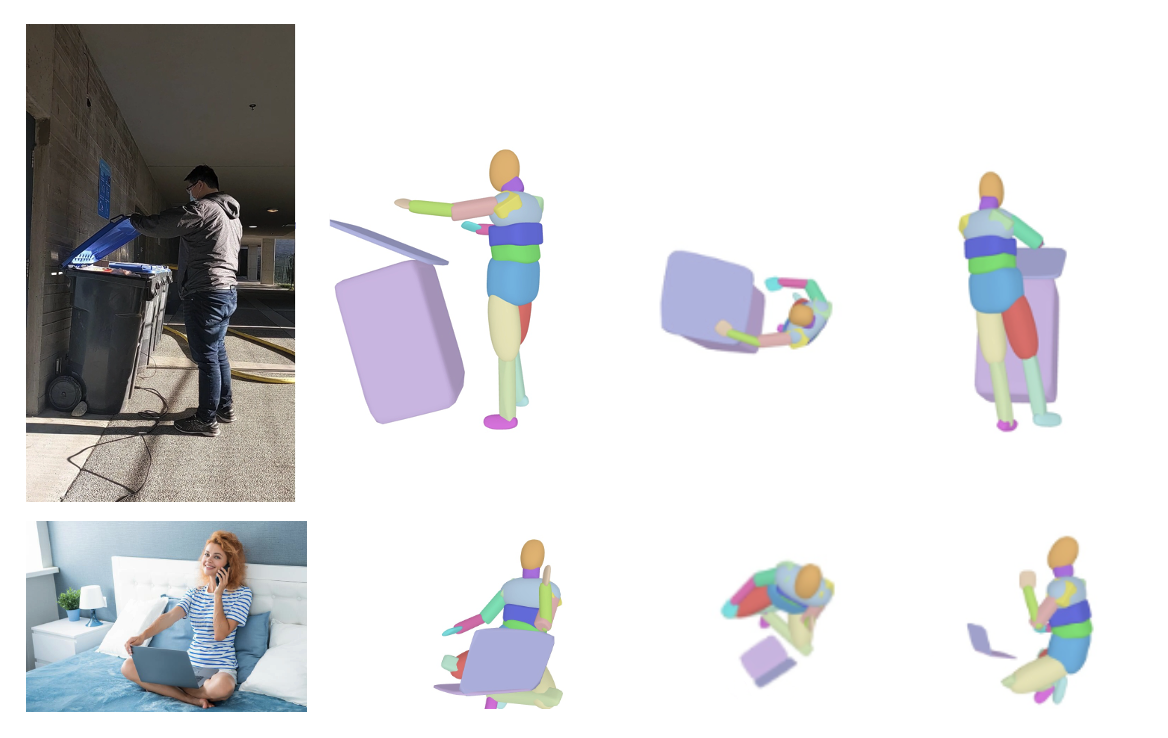}
\vspace{-10px}
\caption{Some failure results.}
\label{fig:limitation}
\vspace{-10px}
\end{figure}

\section{Limitations and Future Work}
Our P3HAOI, the first method for the challenging task of reconstructing primitive-based 3D HOI shapes from single-view images, shows promising results, but limitations exist. 
For example, as shown in Fig.~\ref{fig:limitation}, in some cases, it is hard to predict the right face direction of the object to the human. Some common sense knowledge may be leveraged to overcome this issue. Future work may include more complex articulated objects and more complex situations. We will also apply our dataset to other downstream tasks.

\section{Conclusion}
In this work, we propose a 3D superquadric primitive-based representation and a  3D dataset.
Given this dataset, we propose a task requiring machines to recover the 3D HAOI from images with our superquadric primitive and a baseline. 
We believe our new paradigm would pave a new way for future 3D scene understanding.

\section{Acknowledgement}
This work is supported in part by the
National Natural Science Foundation of China under Grants 62306175, 
National Key R\&D Program of China (No.2021ZD0110704), 
Shanghai Municipal Science and Technology Major Project (2021SHZDZX0102), 
Shanghai Qi Zhi Institute, 
Shanghai Science and Technology Commission (21511101200).

\section{Ethics Statement}
For images with real humans, we will code the faces to protect their portrait rights. For photos collected from the Internet, we will only provide the original links to these photos to protect the copyrighted photographs.

\bibliography{aaai24}

\begin{thebibliography}{48}
\providecommand{\natexlab}[1]{#1}

\bibitem[{Bhatnagar et~al.(2022)Bhatnagar, Xie, Petrov, Sminchisescu, Theobalt,
  and Pons-Moll}]{bhatnagar2022behave}
Bhatnagar, B.~L.; Xie, X.; Petrov, I.~A.; Sminchisescu, C.; Theobalt, C.; and
  Pons-Moll, G. 2022.
\newblock Behave: Dataset and method for tracking human object interactions.
\newblock In \emph{Proceedings of the IEEE/CVF Conference on Computer Vision
  and Pattern Recognition}, 15935--15946.

\bibitem[{Cao et~al.(2017)Cao, Simon, Wei, and Sheikh}]{cao2017realtime}
Cao, Z.; Simon, T.; Wei, S.-E.; and Sheikh, Y. 2017.
\newblock Realtime Multi-Person 2D Pose Estimation using Part Affinity Fields.
\newblock In \emph{CVPR}.

\bibitem[{Dwivedi et~al.(2022)Dwivedi, Athanasiou, Kocabas, and
  Black}]{dwivedi2022learning}
Dwivedi, S.~K.; Athanasiou, N.; Kocabas, M.; and Black, M.~J. 2022.
\newblock Learning to Regress Bodies from Images using Differentiable Semantic
  Rendering.
\newblock arXiv:2110.03480.

\bibitem[{Fang et~al.(2022)Fang, Li, Tang, Xu, Zhu, Xiu, Li, and
  Lu}]{alphapose}
Fang, H.-S.; Li, J.; Tang, H.; Xu, C.; Zhu, H.; Xiu, Y.; Li, Y.-L.; and Lu, C.
  2022.
\newblock AlphaPose: Whole-Body Regional Multi-Person Pose Estimation and
  Tracking in Real-Time.
\newblock \emph{IEEE Transactions on Pattern Analysis and Machine
  Intelligence}.

\bibitem[{Fang et~al.(2017)Fang, Xie, Tai, and Lu}]{fang2017rmpe}
Fang, H.-S.; Xie, S.; Tai, Y.-W.; and Lu, C. 2017.
\newblock {RMPE}: Regional Multi-person Pose Estimation.
\newblock In \emph{ICCV}.

\bibitem[{Gavrila(2000)}]{DBLP:conf/eccv/Gavrila00}
Gavrila, D. 2000.
\newblock Pedestrian Detection from a Moving Vehicle.
\newblock In Vernon, D., ed., \emph{Computer Vision - {ECCV} 2000, 6th European
  Conference on Computer Vision, Dublin, Ireland, June 26 - July 1, 2000,
  Proceedings, Part {II}}, volume 1843 of \emph{Lecture Notes in Computer
  Science}, 37--49. Springer.

\bibitem[{Han et~al.(2021)Han, Gu, Mo, Yi, Hu, Chen, and
  Su}]{han2021compositionally}
Han, S.; Gu, J.; Mo, K.; Yi, L.; Hu, S.; Chen, X.; and Su, H. 2021.
\newblock Compositionally Generalizable 3D Structure Prediction.
\newblock arXiv:2012.02493.

\bibitem[{He et~al.(2021)He, Zhou, Wan, and He}]{he2021single}
He, Q.; Zhou, D.; Wan, B.; and He, X. 2021.
\newblock Single Image 3D Object Estimation with Primitive Graph Networks.
\newblock arXiv:2109.04153.

\bibitem[{Huang et~al.(2022)Huang, Tehari, Black, and
  Tzionas}]{huang2022intercap}
Huang, Y.; Tehari, O.; Black, M.~J.; and Tzionas, D. 2022.
\newblock InterCap: Joint Markerless 3D Tracking of Humans and Objects in
  Interaction.
\newblock arXiv:2209.12354.

\bibitem[{Huang et~al.(2023)Huang, Jampani, Thai, Li, Stojanov, and
  Rehg}]{huang2023shapeclipper}
Huang, Z.; Jampani, V.; Thai, A.; Li, Y.; Stojanov, S.; and Rehg, J.~M. 2023.
\newblock ShapeClipper: Scalable 3D Shape Learning from Single-View Images via
  Geometric and CLIP-based Consistency.
\newblock arXiv:2304.06247.

\bibitem[{Jiang et~al.(2020)Jiang, Kolotouros, Pavlakos, Zhou, and
  Daniilidis}]{DBLP:journals/corr/abs-2006-08586}
Jiang, W.; Kolotouros, N.; Pavlakos, G.; Zhou, X.; and Daniilidis, K. 2020.
\newblock Coherent Reconstruction of Multiple Humans from a Single Image.
\newblock \emph{CoRR}, abs/2006.08586.

\bibitem[{Joo, Neverova, and Vedaldi(2020)}]{joo2020eft}
Joo, H.; Neverova, N.; and Vedaldi, A. 2020.
\newblock Exemplar Fine-Tuning for 3D Human Pose Fitting Towards In-the-Wild 3D
  Human Pose Estimation.
\newblock In \emph{3DV}.

\bibitem[{Kluger et~al.(2021)Kluger, Ackermann, Brachmann, Yang, and
  Rosenhahn}]{kluger2021cuboids}
Kluger, F.; Ackermann, H.; Brachmann, E.; Yang, M.~Y.; and Rosenhahn, B. 2021.
\newblock Cuboids Revisited: Learning Robust 3D Shape Fitting to Single RGB
  Images.
\newblock arXiv:2105.02047.

\bibitem[{Laine et~al.(2020)Laine, Hellsten, Karras, Seol, Lehtinen, and
  Aila}]{Laine2020diffrast}
Laine, S.; Hellsten, J.; Karras, T.; Seol, Y.; Lehtinen, J.; and Aila, T. 2020.
\newblock Modular Primitives for High-Performance Differentiable Rendering.
\newblock \emph{ACM Transactions on Graphics}, 39(6).

\bibitem[{Li et~al.(2019{\natexlab{a}})Li, Wang, Zhu, Mao, Fang, and
  Lu}]{li2019crowdpose}
Li, J.; Wang, C.; Zhu, H.; Mao, Y.; Fang, H.-S.; and Lu, C. 2019{\natexlab{a}}.
\newblock Crowdpose: Efficient crowded scenes pose estimation and a new
  benchmark.
\newblock In \emph{Proceedings of the IEEE/CVF conference on computer vision
  and pattern recognition}, 10863--10872.

\bibitem[{Li et~al.(2020)Li, Liu, Lu, Wang, Liu, Li, and Lu}]{li2020detailed}
Li, Y.-L.; Liu, X.; Lu, H.; Wang, S.; Liu, J.; Li, J.; and Lu, C. 2020.
\newblock Detailed 2D-3D Joint Representation for Human-Object Interaction.
\newblock arXiv:2004.08154.

\bibitem[{Li et~al.(2019{\natexlab{b}})Li, Zhou, Huang, Xu, Ma, Fang, Wang, and
  Lu}]{li2019transferable}
Li, Y.-L.; Zhou, S.; Huang, X.; Xu, L.; Ma, Z.; Fang, H.-S.; Wang, Y.; and Lu,
  C. 2019{\natexlab{b}}.
\newblock Transferable Interactiveness Knowledge for Human-Object Interaction
  Detection.
\newblock In \emph{Proceedings of the IEEE Conference on Computer Vision and
  Pattern Recognition}, 3585--3594.

\bibitem[{Liu et~al.(2022{\natexlab{a}})Liu, Xu, Fu, Qian, Yu, Han, and
  Lu}]{liu2022akb}
Liu, L.; Xu, W.; Fu, H.; Qian, S.; Yu, Q.; Han, Y.; and Lu, C.
  2022{\natexlab{a}}.
\newblock AKB-48: A Real-World Articulated Object Knowledge Base.
\newblock In \emph{Proceedings of the IEEE/CVF Conference on Computer Vision
  and Pattern Recognition}, 14809--14818.

\bibitem[{Liu et~al.(2022{\natexlab{b}})Liu, Li, Wu, Tai, Lu, and
  Tang}]{liu2022interactiveness}
Liu, X.; Li, Y.-L.; Wu, X.; Tai, Y.-W.; Lu, C.; and Tang, C.-K.
  2022{\natexlab{b}}.
\newblock Interactiveness Field in Human-Object Interactions.
\newblock In \emph{Proceedings of the IEEE/CVF Conference on Computer Vision
  and Pattern Recognition}, 20113--20122.

\bibitem[{Loper et~al.(2015)Loper, Mahmood, Romero, Pons-Moll, and
  Black}]{SMPL:2015}
Loper, M.; Mahmood, N.; Romero, J.; Pons-Moll, G.; and Black, M.~J. 2015.
\newblock {SMPL}: A Skinned Multi-Person Linear Model.
\newblock \emph{ACM Trans. Graphics (Proc. SIGGRAPH Asia)}, 34(6):
  248:1--248:16.

\bibitem[{Oquab et~al.(2023)Oquab, Darcet, Moutakanni, Vo, Szafraniec,
  Khalidov, Fernandez, Haziza, Massa, El-Nouby, Howes, Huang, Xu, Sharma, Li,
  Galuba, Rabbat, Assran, Ballas, Synnaeve, Misra, Jegou, Mairal, Labatut,
  Joulin, and Bojanowski}]{oquab2023dinov2}
Oquab, M.; Darcet, T.; Moutakanni, T.; Vo, H.~V.; Szafraniec, M.; Khalidov, V.;
  Fernandez, P.; Haziza, D.; Massa, F.; El-Nouby, A.; Howes, R.; Huang, P.-Y.;
  Xu, H.; Sharma, V.; Li, S.-W.; Galuba, W.; Rabbat, M.; Assran, M.; Ballas,
  N.; Synnaeve, G.; Misra, I.; Jegou, H.; Mairal, J.; Labatut, P.; Joulin, A.;
  and Bojanowski, P. 2023.
\newblock DINOv2: Learning Robust Visual Features without Supervision.

\bibitem[{Park et~al.(2019)Park, Florence, Straub, Newcombe, and
  Lovegrove}]{park2019deepsdf}
Park, J.~J.; Florence, P.; Straub, J.; Newcombe, R.; and Lovegrove, S. 2019.
\newblock DeepSDF: Learning Continuous Signed Distance Functions for Shape
  Representation.
\newblock arXiv:1901.05103.

\bibitem[{Paschalidou et~al.(2021)Paschalidou, Katharopoulos, Geiger, and
  Fidler}]{paschalidou2021neural}
Paschalidou, D.; Katharopoulos, A.; Geiger, A.; and Fidler, S. 2021.
\newblock Neural Parts: Learning Expressive 3D Shape Abstractions with
  Invertible Neural Networks.
\newblock arXiv:2103.10429.

\bibitem[{Rueegg et~al.(2022)Rueegg, Zuffi, Schindler, and
  Black}]{rueegg2022barc}
Rueegg, N.; Zuffi, S.; Schindler, K.; and Black, M.~J. 2022.
\newblock BARC: Learning to Regress 3D Dog Shape from Images by Exploiting
  Breed Information.
\newblock arXiv:2203.15536.

\bibitem[{Sitzmann, Zollhöfer, and Wetzstein(2020)}]{sitzmann2020scene}
Sitzmann, V.; Zollhöfer, M.; and Wetzstein, G. 2020.
\newblock Scene Representation Networks: Continuous 3D-Structure-Aware Neural
  Scene Representations.
\newblock arXiv:1906.01618.

\bibitem[{Sun et~al.(2021{\natexlab{a}})Sun, Bao, Liu, Fu, Black, and
  Mei}]{sun2021monocular}
Sun, Y.; Bao, Q.; Liu, W.; Fu, Y.; Black, M.~J.; and Mei, T.
  2021{\natexlab{a}}.
\newblock Monocular, one-stage, regression of multiple 3d people.
\newblock In \emph{Proceedings of the IEEE/CVF International Conference on
  Computer Vision}, 11179--11188.

\bibitem[{Sun et~al.(2021{\natexlab{b}})Sun, Bao, Liu, Fu, Michael~J., and
  Mei}]{ROMP}
Sun, Y.; Bao, Q.; Liu, W.; Fu, Y.; Michael~J., B.; and Mei, T.
  2021{\natexlab{b}}.
\newblock Monocular, One-stage, Regression of Multiple 3D People.
\newblock In \emph{ICCV}.

\bibitem[{Wang et~al.(2022)Wang, Li, Kuo, Kocabas, Aksan, and
  Hilliges}]{wang2022reconstruction}
Wang, X.; Li, G.; Kuo, Y.-L.; Kocabas, M.; Aksan, E.; and Hilliges, O. 2022.
\newblock Reconstructing Action-Conditioned Human-Object Interactions Using
  Commonsense Knowledge Priors.
\newblock In \emph{International Conference on 3D Vision (3DV)}.

\bibitem[{Wu et~al.(2017)Wu, Zhang, Xue, Freeman, and
  Tenenbaum}]{wu2017learning}
Wu, J.; Zhang, C.; Xue, T.; Freeman, W.~T.; and Tenenbaum, J.~B. 2017.
\newblock Learning a Probabilistic Latent Space of Object Shapes via 3D
  Generative-Adversarial Modeling.
\newblock arXiv:1610.07584.

\bibitem[{Wu et~al.(2023)Wu, Li, Jakab, Rupprecht, and
  Vedaldi}]{wu2023magicpony}
Wu, S.; Li, R.; Jakab, T.; Rupprecht, C.; and Vedaldi, A. 2023.
\newblock {MagicPony}: Learning Articulated 3D Animals in the Wild.

\bibitem[{Wu et~al.(2022)Wu, Li, Liu, Zhang, Wu, and Lu}]{wu2022mining}
Wu, X.; Li, Y.-L.; Liu, X.; Zhang, J.; Wu, Y.; and Lu, C. 2022.
\newblock Mining Cross-Person Cues for Body-Part Interactiveness Learning in
  HOI Detection.
\newblock arXiv:2207.14192.

\bibitem[{Wu et~al.(2019)Wu, Kirillov, Massa, Lo, and
  Girshick}]{wu2019detectron2}
Wu, Y.; Kirillov, A.; Massa, F.; Lo, W.-Y.; and Girshick, R. 2019.
\newblock Detectron2.
\newblock \url{https://github.com/facebookresearch/detectron2}.

\bibitem[{Xiang et~al.(2020)Xiang, Qin, Mo, Xia, Zhu, Liu, Liu, Jiang, Yuan,
  Wang et~al.}]{xiang2020sapien}
Xiang, F.; Qin, Y.; Mo, K.; Xia, Y.; Zhu, H.; Liu, F.; Liu, M.; Jiang, H.;
  Yuan, Y.; Wang, H.; et~al. 2020.
\newblock Sapien: A simulated part-based interactive environment.
\newblock In \emph{Proceedings of the IEEE/CVF Conference on Computer Vision
  and Pattern Recognition}, 11097--11107.

\bibitem[{Xie, Bhatnagar, and Pons-Moll(2022)}]{xie2022chore}
Xie, X.; Bhatnagar, B.~L.; and Pons-Moll, G. 2022.
\newblock CHORE: Contact, Human and Object REconstruction from a single RGB
  image.
\newblock In \emph{European Conference on Computer Vision ({ECCV})}.
  {Springer}.

\bibitem[{Xie, Bhatnagar, and Pons-Moll(2023)}]{xie2023vistracker}
Xie, X.; Bhatnagar, B.~L.; and Pons-Moll, G. 2023.
\newblock Visibility Aware Human-Object Interaction Tracking from Single RGB
  Camera.
\newblock In \emph{IEEE Conference on Computer Vision and Pattern Recognition
  (CVPR)}.

\bibitem[{Xu et~al.(2021)Xu, Joo, Mori, and Savva}]{xu2021d3dhoi}
Xu, X.; Joo, H.; Mori, G.; and Savva, M. 2021.
\newblock D3D-HOI: Dynamic 3D Human-Object Interactions from Videos.
\newblock \emph{arXiv preprint arXiv:2108.08420}.

\bibitem[{Yao et~al.(2021)Yao, Hung, Jampani, and Yang}]{yao2021discovering}
Yao, C.-H.; Hung, W.-C.; Jampani, V.; and Yang, M.-H. 2021.
\newblock Discovering 3D Parts from Image Collections.
\newblock arXiv:2107.13629.

\bibitem[{Yao et~al.(2022)Yao, Hung, Li, Rubinstein, Yang, and
  Jampani}]{yao2022lassie}
Yao, C.-H.; Hung, W.-C.; Li, Y.; Rubinstein, M.; Yang, M.-H.; and Jampani, V.
  2022.
\newblock LASSIE: Learning Articulated Shapes from Sparse Image Ensemble via 3D
  Part Discovery.
\newblock \emph{arXiv preprint arXiv:2207.03434}.

\bibitem[{Yao et~al.(2023{\natexlab{a}})Yao, Hung, Li, Rubinstein, Yang, and
  Jampani}]{yao2023hilassie}
Yao, C.-H.; Hung, W.-C.; Li, Y.; Rubinstein, M.; Yang, M.-H.; and Jampani, V.
  2023{\natexlab{a}}.
\newblock Hi-LASSIE: High-Fidelity Articulated Shape and Skeleton Discovery
  from Sparse Image Ensemble.
\newblock arXiv:2212.11042.

\bibitem[{Yao et~al.(2023{\natexlab{b}})Yao, Raj, Hung, Li, Rubinstein, Yang,
  and Jampani}]{yao2023artic3d}
Yao, C.-H.; Raj, A.; Hung, W.-C.; Li, Y.; Rubinstein, M.; Yang, M.-H.; and
  Jampani, V. 2023{\natexlab{b}}.
\newblock ARTIC3D: Learning Robust Articulated 3D Shapes from Noisy Web Image
  Collections.
\newblock arXiv:2306.04619.

\bibitem[{Yavartanoo et~al.(2021)Yavartanoo, Chung, Neshatavar, and
  Lee}]{yavartanoo20213dias}
Yavartanoo, M.; Chung, J.; Neshatavar, R.; and Lee, K.~M. 2021.
\newblock 3DIAS: 3D Shape Reconstruction with Implicit Algebraic Surfaces.
\newblock arXiv:2108.08653.

\bibitem[{Zhang et~al.(2021{\natexlab{a}})Zhang, Liao, Liu, Lu, Wang, Gao, and
  Li}]{zhang2021mining}
Zhang, A.; Liao, Y.; Liu, S.; Lu, M.; Wang, Y.; Gao, C.; and Li, X.
  2021{\natexlab{a}}.
\newblock Mining the Benefits of Two-stage and One-stage HOI Detection.
\newblock arXiv:2108.05077.

\bibitem[{Zhang et~al.(2023)Zhang, Tian, Zhang, Li, An, Sun, and
  Liu}]{pymafx2023}
Zhang, H.; Tian, Y.; Zhang, Y.; Li, M.; An, L.; Sun, Z.; and Liu, Y. 2023.
\newblock PyMAF-X: Towards Well-aligned Full-body Model Regression from
  Monocular Images.
\newblock \emph{IEEE Transactions on Pattern Analysis and Machine
  Intelligence}.

\bibitem[{Zhang et~al.(2021{\natexlab{b}})Zhang, Tian, Zhou, Ouyang, Liu, Wang,
  and Sun}]{pymaf2021}
Zhang, H.; Tian, Y.; Zhou, X.; Ouyang, W.; Liu, Y.; Wang, L.; and Sun, Z.
  2021{\natexlab{b}}.
\newblock PyMAF: 3D Human Pose and Shape Regression with Pyramidal Mesh
  Alignment Feedback Loop.
\newblock In \emph{Proceedings of the IEEE International Conference on Computer
  Vision}.

\bibitem[{Zhang et~al.(2020)Zhang, Pepose, Joo, Ramanan, Malik, and
  Kanazawa}]{zhang2020perceiving}
Zhang, J.~Y.; Pepose, S.; Joo, H.; Ramanan, D.; Malik, J.; and Kanazawa, A.
  2020.
\newblock Perceiving 3D Human-Object Spatial Arrangements from a Single Image
  in the Wild.
\newblock arXiv:2007.15649.

\bibitem[{Zhou et~al.(2020)Zhou, Wu, Li, Cao, Ye, Saragih, Li, and
  Sheikh}]{DBLP:journals/corr/abs-2006-04325}
Zhou, Y.; Wu, C.; Li, Z.; Cao, C.; Ye, Y.; Saragih, J.~M.; Li, H.; and Sheikh,
  Y. 2020.
\newblock Fully Convolutional Mesh Autoencoder using Efficient Spatially
  Varying Kernels.
\newblock \emph{CoRR}, abs/2006.04325.

\bibitem[{Zuffi et~al.(2019)Zuffi, Kanazawa, Berger-Wolf, and
  Black}]{zuffi2019threed}
Zuffi, S.; Kanazawa, A.; Berger-Wolf, T.; and Black, M.~J. 2019.
\newblock Three-D Safari: Learning to Estimate Zebra Pose, Shape, and Texture
  from Images "In the Wild".
\newblock arXiv:1908.07201.

\bibitem[{Zuffi, Kanazawa, and Black(2018)}]{8578514}
Zuffi, S.; Kanazawa, A.; and Black, M.~J. 2018.
\newblock Lions and Tigers and Bears: Capturing Non-rigid, 3D, Articulated
  Shape from Images.
\newblock In \emph{2018 IEEE/CVF Conference on Computer Vision and Pattern
  Recognition}, 3955--3963.

\end{thebibliography}

\end{document}